\documentclass{article}

\usepackage{arxiv}

\usepackage[utf8]{inputenc} 
\usepackage[T1]{fontenc}    
\usepackage{hyperref}       
\usepackage{url}            
\usepackage{booktabs}       
\usepackage{amsfonts}       
\usepackage{nicefrac}       
\usepackage{microtype}      
\usepackage{lipsum}
\usepackage{graphicx}
\graphicspath{ {./images/} }
\usepackage{amsmath}
\usepackage{amssymb}
\usepackage{enumitem}
\setlist[itemize,enumerate]{topsep=0pt, itemsep=0pt, partopsep=0pt, parsep=0pt}
\usepackage{multirow}
\usepackage{gensymb}

\title{Training-free zero-shot 3D symmetry detection with visual features back-projected to geometry}

\author{
Isaac Aguirre\\
Department of Computer Science\\
 University of Chile\\
 Santiago, Chile\\
\texttt{iaguirre@dcc.uchile.cl}
\And
Ivan Sipiran\\
Department of Computer Science\\
 University of Chile\\
 Santiago, Chile\\
\texttt{isipiran@dcc.uchile.cl}
}

\begin{document}
\maketitle
\begin{abstract}
We present a simple yet effective training-free approach for zero-shot 3D symmetry detection that leverages visual features from foundation vision models such as DINOv2. Our method extracts features from rendered views of 3D objects and backprojects them onto the original geometry. We demonstrate the symmetric invariance of these features and use them to identify reflection-symmetry planes through a proposed algorithm. Experiments on a subset of ShapeNet demonstrate that our approach outperforms both traditional geometric methods and learning-based approaches without requiring any training data. Our work demonstrates how foundation vision models can help in solving complex 3D geometric problems such as symmetry detection.
\end{abstract}


\section{Introduction}
\label{sec:intro}
Computer vision and deep learning have revolutionized the way machines interpret visual data, enabling applications from autonomous driving to cultural heritage preservation. In the analysis of 3D objects, surfaces of the shape are encoded as a vector of numbers, often called characteristics, descriptors, or features, that capture local geometric patterns and textures much like humans perceive shapes and symmetries by eye.

Among these tasks, symmetry detection is important in reconstruction, recognition, and compression of 3D models. Symmetries not only reveal underlying regularities in both natural and man made objects but also serves as a guide for robust surface completion algorithms. Existing symmetry detection methodologies are based on handcrafted or learned features that capture only a fraction of an object's surface, leaving behind rich semantic, textural, or global shape information.

Feature backprojection procedures produces high quality 3D features, by rendering a 3D mesh or point cloud from multiple viewpoints, extracting image based characteristics via foundation vision models such as DINOv2, and mapping those characteristics back onto the 3D geometry, resulting in robust and information rich features representations. However, current feature backprojection procedures have not been fully adapted or leverage this powerful back-projected features to focus on symmetry detection.

In this paper, we propose a novel zero-shot approach for 3D symmetry detection that requires no training data or parameter tuning. Using these back-projected features and analyzing their distribution, we can identify symmetry planes through a proposed algorithm. We demonstrate that these features exhibit remarkable invariance to symmetric transformations, allowing us to identify symmetric point correspondences directly in feature space. From these correspondences, we derive candidate symmetry planes and rank them based on their geometric consistency.

To summarize, our \textbf{main contributions} are as follows:
\begin{enumerate}
    \item We present the first training-free approach for 3D symmetry detection that leverages visual features from foundation vision models,  eliminating the need for training data or fine-tuning.
    \item We demonstrate that visual features back-projected from foundation vision models are highly invariant to symmetry transformations under certain symmetric-beneficial conditions, providing empirical validation for this.
    \item We develop an effective algorithm for detecting symmetry planes from feature correspondences, incorporating geometric constraints and confidence scoring to identify true symmetries.
    \item We achieve state-of-the-art performance on a standard experiment, surpassing both traditional geometric methods and specialized learning-based approaches, with a $\sim15\%$ improvement in F-score compared to previous best results.
 \end{enumerate}

\begin{figure}[ht]
\begin{center}
\fbox{\includegraphics[width=.9\linewidth]{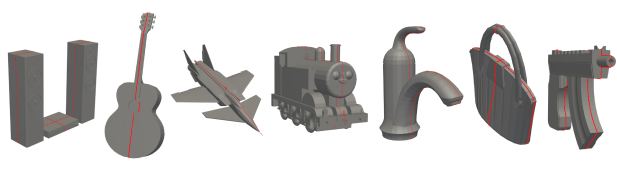}}
\end{center}
   \caption{Different 3D objects and symmetry planes detected by our method. Symmetry planes are outlined in red.}
\label{fig:intro_figure}
\end{figure}


\section{Related Works}
\label{Sec:Related}
The problem of symmetry detection in 3D objects has been extensively studied in computer vision and graphics. We organize relevant prior work into three main categories: traditional geometric approaches, learning-based methods, and foundation vision model related techniques.

\vspace{0.1em}\noindent\textbf{Geometric Approaches}
These approaches rely on analyzing intrinsic object properties. Principal Component Analysis (PCA) identifies potential symmetry planes through principal axes analysis but fails with non-aligned symmetries. More sophisticated methods include Oriented Bounding Box (OBB)~\cite{Chang2011}, Reflective Symmetry Descriptor (RSD)~\cite{Kazhdan2002} using voxel-based representation, Approximate Discrete Symmetry (ADS)~\cite{Martinet2006} matching surface points by geometric properties, and Planar-Reflective Symmetry Transform (PRST)~\cite{Podolak2006}, enhanced by Gaussian Euclidean Distance Transform (GEDT)~\cite{Podolak2006}. Methods like Partial and Approximate Symmetry Detection (PASD)~\cite{Mitra2006} and Probably Approximately Symmetric (PAS)~\cite{Korman2015} address approximately symmetric or incomplete objects. Diffusion-based methods (Diffussion)~\cite{Sipiran2014} capture symmetries by analyzing how symmetry-aware descriptors propagate across the object's surface. Despite their mathematical elegance, purely geometric approaches struggle with complex geometries, noise, or missing data.

\vspace{0.1em}\noindent\textbf{Learning-based Methods}
These methods leverage neural networks to improve detection accuracy. PRS-Net~\cite{Gao2021} pioneered unsupervised neural networks for planar symmetry detection. E3Sym~\cite{Li2023} introduced equivariant neural networks respecting the E(3) symmetry group, while Self-Sym~\cite{Aguirre2025} proposed a dataset-free approach using self-supervised learning. Though impressive, these methods require significant computational resources and often don't generalize well to unseen objects during training.

\vspace{0.1em}\noindent\textbf{Hybrid Methods}
Langevin~\cite{Langevin2024} uses geometric and generative models alike, combining voting-based detection with Riemannian Langevin Dynamics for robust, training-free symmetry detection.

\vspace{0.1em}\noindent\textbf{Feature backprojection}
There are techniques that render 3D objects from multiple viewpoints and map extracted visual features back to the geometry. Foundation vision models like DINOv2~\cite{Oquab2023} or CLIP~\cite{CLIP} generate rich visual features for 2D images. Works like Back-to-3D (BT3D)~\cite{Wimmer2024} and Comprehensive model for Parts Segmentation (COPS)~\cite{garosi2025} utilize feature backprojection for 3D meshes and point clouds respectively, but haven't fully exploited these features for symmetry detection.

\vspace{0.1em}\noindent\textbf{Link Between Feature Backprojection and Symmetry Detection}
Backprojected features are promising for symmetry detection because they: (1) encode both local geometric details and global contextual information, (2) identify semantic similarities between visually related elements, and (3) create comprehensive representations by utilizing features from multiple viewpoints. Our work extends this research by specifically adapting foundation vision model features for symmetry detection.

\section{Our Method}
\label{Sec:Method}
Our approach consists of two main phases: computing object descriptors and detecting symmetries. As previously mentioned, our method does not require any training and does not rely on ground truth data, except for evaluation purposes. In this section, we provide a detailed explanation of each phase.

\subsection{Feature Extraction}
Our approach leverages visual features extracted from foundation vision models and projects them onto 3D geometry to identify symmetry patterns. While prior works have demonstrated the effectiveness of feature backprojection for various 3D tasks such as keypoint detection~\cite{Wimmer2024} and segmentation~\cite{garosi2025}, we optimize this process specifically for symmetry detection. Our feature extraction pipeline consists of four key steps:

\vspace{0.1em}\noindent\textbf{Multi-view Rendering}: We generate renders ranging from $[6, 14, 26, 42, 62, 86, 144]$ viewpoints (more on Section \ref{Sec:Invariance}) sampled using the Fibonacci spiral method which provides more uniform coverage of the viewing sphere compared to regular sphere sampling~\cite{Fibonacci2009}. For each viewpoint, we place a virtual camera and render the object using flat illumination model with a uniform gray material to ensure consistent lighting conditions, finally, following Self-Sym~\cite{Aguirre2025}, we augment each viewpoint with four rotated renders (0°, 90°, 180°, and 270°).

\vspace{0.1em}\noindent\textbf{Feature Extraction}: Each rendered image is processed using DINOv2 with registers~\cite{Darcet2023} in its small variant, selected for its robustness to noise and high-frequency content. The transformer-based architecture of DINOv2 produces patch-level features of dimension 384 (preserved throughout the entire process) that capture both local geometric details and global contextual information.

\vspace{0.1em}\noindent\textbf{Feature Backprojection}: For each rendered viewpoint where a mesh vertex is visible, that is, directly visible in the render given by pixel-fragment information or contiguous to a visible mesh given by the former, we determine its corresponding pixel location in the rendered image. We then assign the feature vector from that pixel to the vertex. When a vertex is visible from multiple viewpoints, we aggregate the features by taking their average across all viewpoints where the vertex is visible. This process produce a feature set $F_V = \{f_{v_1}, f_{v_2}, \ldots, f_{v_m}\}$, where each vertex $v_i$ of the original mesh has a corresponding feature vector $f_{v_i}$ of dimension $384$.

\vspace{0.1em}\noindent\textbf{Feature-mesh Sampling}: To ensure consistent feature distribution regardless of the original mesh topology, we sample 10,000 points $P = \{p_1, p_2, \ldots, p_n\}$ uniformly across the object's surface. For each sampled point $p_i$ located on a mesh face with vertices $(v_a, v_b, v_c)$ and barycentric coordinates $(\alpha, \beta, \gamma)$ such that $p_i = \alpha v_a + \beta v_b + \gamma v_c$ where $\alpha + \beta + \gamma = 1$, we interpolate its feature vector as $f_i = \alpha f_{v_a} + \beta f_{v_b} + \gamma f_{v_c}$.

This interpolation ensures that our final feature set $F(P) = \{f_1, f_2, \ldots, f_n\}$ preserves the rich semantic information captured by the foundation model while providing uniform spatial distribution across the object's surface. 

Each of our key design choices such as viewpoint sampling, image rotation, point sampling and interpolation of features are justified in Section \ref{Sec:Invariance} and serve as the foundation for our symmetry detection algorithm, described in the following subsection.

\subsection{Symmetry Detection}
\label{Subsec:Symmetry_det}
At this stage, the input object is represented as a set of points with corresponding feature descriptors, specifically a point set $P = \{p_1, p_2, \ldots, p_n\}$ and a feature set $F(P) = \{f_1, f_2, \ldots, f_n\}$, where $f_i = F(p_i)$ is the feature descriptor corresponding to point $p_i$. We now describe the steps involved in computing symmetry planes.
    
\vspace{0.1em}\noindent\textbf{Symmetric Point Matching} Our hypothesis is that the extracted features are invariant to the object's symmetries. Therefore, our first objective is to identify potential symmetric correspondences to facilitate symmetry detection. For each point $p_i$ in the object, we compute the two nearest neighbors $p_j$ and $p_k$ in feature space, using the L1 or Manhattan distance. Specifically, we find points $p_j$ and $p_k$ such that:
        
\begin{equation}
    d_F(p_i, p_j) \leq d_F(p_i, p_k) \leq d_F(p_i, p_l) \quad \forall l \neq i,j,k
\end{equation}

\noindent Where the feature-space distance $d_F$ is defined as:

\begin{equation}
    d_F(p_i, p_l) = \|F(p_i) - F(p_l)\|_1
\end{equation}

\noindent The output of this step is a set of point trios $\{(p_i, p_j, p_k)\}$ that are potential symmetric correspondences.

\vspace{0.1em}\noindent\textbf{Candidate Plane Computation} For each trio of points $(p_i, p_j, p_k)$, we compute four candidate symmetry planes: 

\noindent Three planes from each possible pair of points within the trio. For a pair of points $(p_a, p_b)$ (where $a,b \in \{i,j,k\}$ and $a \neq b$), the candidate symmetry plane is defined by the normal vector $\vec{n}_{ab}$ and a point $c_{ab}$ on the plane:
\begin{equation}
    \vec{n}_{ab} = \frac{p_b - p_a}{\|p_b - p_a\|_2},\qquad
    c_{ab} = \frac{p_a + p_b}{2}
\end{equation}

\noindent One plane from the trio as a whole. For a trio of points $(p_i, p_j, p_k)$, we compute the plane passing through these three points, defined by the normal vector $\vec{n}_{ijk}$ and a point $c_{ijk}$ on the plane:
\begin{equation}
    c_{ijk} = \frac{p_i + p_j + p_k}{3},\qquad
    \vec{n}_{ijk} = \frac{(p_j - p_i) \times (p_k - p_i)}{\|(p_j - p_i) \times (p_k - p_i)\|_2}
\end{equation}

\vspace{0.1em}\noindent\textbf{Candidate Plane Filtering} To identify global symmetry planes, we assume that the 3D object is centered at the origin. We filter candidate planes whose distance from the origin exceeds 5\% of the object's diagonal $O_d$. This threshold is intentionally strict to ensure that only planes passing near the origin are considered, as global symmetry planes for centered objects must pass very close to the object's center. For a given plane with normal $\vec{n}$ and point $c$ on the plane, the following must be satisfied:

\begin{equation}
    |\vec{n} \cdot c| \leq 0.05 \cdot O_d
\end{equation}

\vspace{0.1em}\noindent\textbf{Symmetry Verification} For each of the remaining candidate planes, we assess their quality as global symmetry planes by measuring how well the object maps to itself when reflected across the plane. For a point $p$ and a plane with normal $\vec{n}$ and point $c$, the reflection $p'$ is given by $p' = p - 2(\vec{n} \cdot (p - c))\vec{n}$. We then compute the Chamfer distance between the original point set $P$ and its reflection $P' = \{p'_1, p'_2, \ldots, p'_n\}$:

\begin{equation}
    d_{\text{Chamfer}}(P, P') = \frac{1}{2}\left(\frac{1}{|P|} \sum_{p \in P} \min_{p' \in P'} \|p - p'\|_2^2 + \frac{1}{|P'|} \sum_{p' \in P'} \min_{p \in P} \|p' - p\|_2^2\right)
\end{equation}

\vspace{0.1em}\noindent\textbf{Final Selection} A plane is considered a valid symmetry plane if the Chamfer distance is below $\tau_1 = 0.01$. This threshold was chosen to be strict to ensure that only high-quality symmetries with minimal geometric discrepancy between an object and its reflection are maintained, for it is essential to distinguish true global symmetries from approximate or partial ones. The remaining planes are sorted in ascending order of their Chamfer distance. A final filtering step removes redundant planes by clustering them based on an angular threshold $\tau_2 = 1\degree$. The confidence score for each plane is computed as:
\begin{equation}
    C = 1 - \frac{d_{\text{Chamfer}}}{\tau}
\end{equation}
\noindent At the end of this process, we retain at most $k = 10$ symmetry planes with the highest confidence scores.

\section{Feature Invariance Analysis}
\label{Sec:Invariance}
The effectiveness of our symmetry detection approach relies critically on the invariance property of the back-projected features with respect to symmetric transformations. If features are truly invariant to symmetry, points that are symmetric counterparts of each other should have similar feature representations. This section analyzes this property experimentally and justifies our key design choices.

\subsection{Theoretical Formulation}

For a 3D object $O$ with a reflective symmetry plane $P$, let $T_P$ be the reflection transformation defined by $P$. A feature representation $F$ is considered invariant to the symmetry defined by $P$ if:
\begin{equation}
    F(x) \approx F(T_P(x)) \quad \text{for all } x \in O
\end{equation}

\noindent In practice, perfect invariance is rarely achieved due to various factors, including discretization, rendering artifacts, and the inherent behavior of the foundation model. We therefore quantify the degree of invariance using a feature discrepancy measure. L1 distance is used because is computationally efficient (summing absolute differences reduces costs compared to L2 or cosine similarity) making it ideal for the multiple distance computations required in Subsection \ref{Subsec:Symmetry_det}.
\begin{equation}
    \mathcal{E}(O) = \frac{1}{|O|}\sum_{x \in O} \|F(x) - F(\overline{x})\|_1
\end{equation}

\noindent Where $\overline{x} = \underset{y \in O}{\arg\min} \; \| y - T_P(x) \|_2$ is the nearest point to $T_P(x)$ on the object's surface. Lower values of $\mathcal{E}$ indicate greater feature invariance to symmetry transformations.

\subsection{Experimental Design}
We conducted a series of experiments to analyze feature invariance and determine optimal parameters for our approach. For each experiment, we selected a fixed subset of 50 objects from the ShapeNet dataset with known ground-truth symmetry planes defined in the Subsection \ref{Subsec:Dataset}. We computed the feature discrepancy measure $\mathcal{E}$ for each object using different configurations and averaged the results across all 50 objects. Then we investigated the following factors:
\begin{enumerate}
    \item Viewpoint sampling: Regular sphere sampling vs. Fibonacci spiral sampling. We use 6, 14, 26, 42, 62, 86, and 114 viewpoints because these correspond to partition levels 1 through 7 in regular sphere sampling. While regular sampling requires specific numbers of views due to its partitioning structure, Fibonacci spiral sampling can use any number of viewpoints since it does not rely on partitions
    \item Sampling density for Feature-mesh sampling: 1K, 3K, 5K, 7K, and 10K points
    \item Image rotation augmentation: 1 (0°), 2 (0°, 180°), 3 (0°, 90°, 270°), and 4 (0°, 90°, 180°, 270°) rotations and a control condition repeating the same image 4 times without rotation (T4)
    \item Random pairing: We include random pairing for comparison with our proposed Feature Invariance pairing
\end{enumerate}

\begin{figure}
\centering
\begin{tabular}{cc}
\hbox{\fbox{\includegraphics[width=0.45\linewidth]{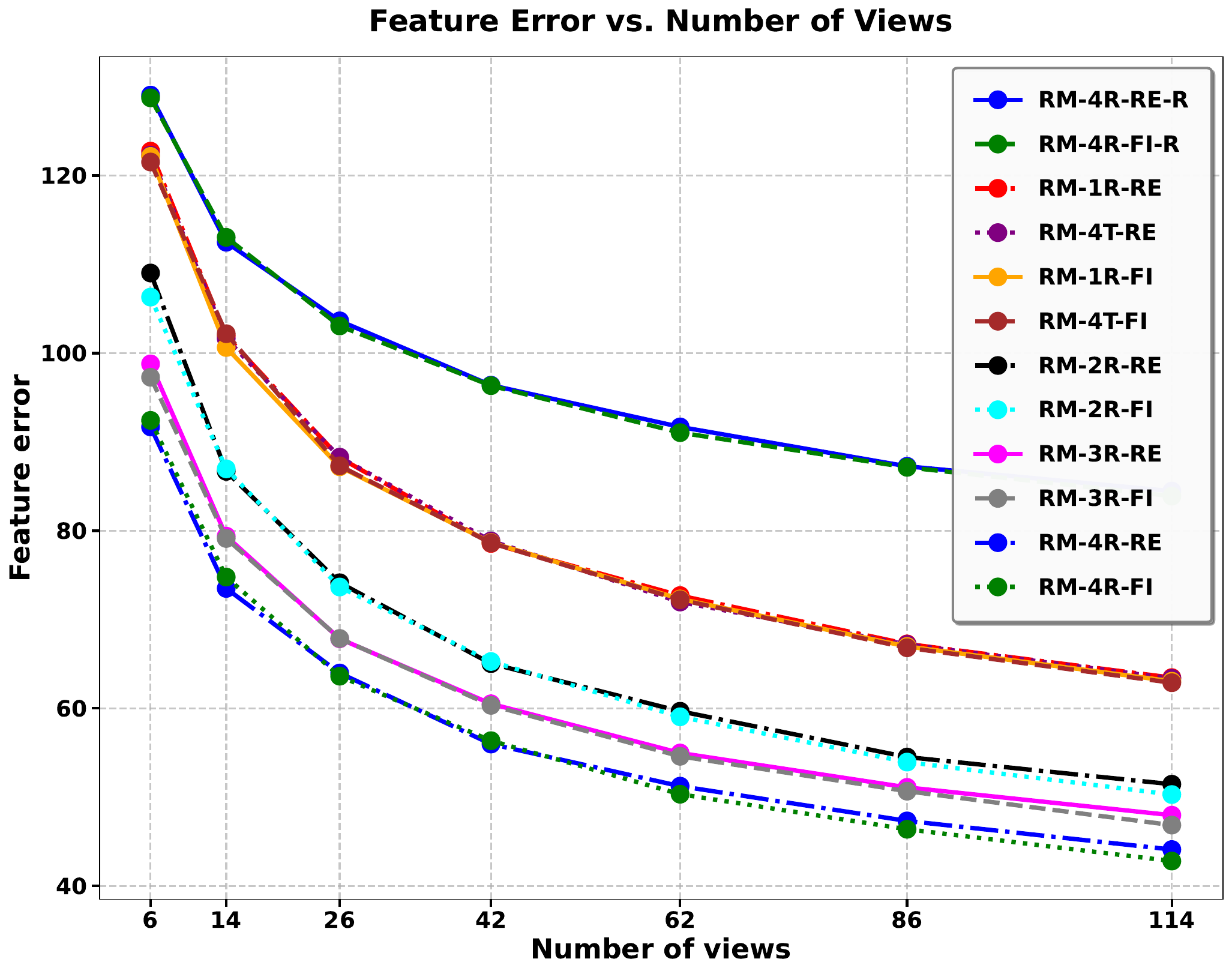}}}&
\hbox{\fbox{\includegraphics[width=0.45\linewidth]{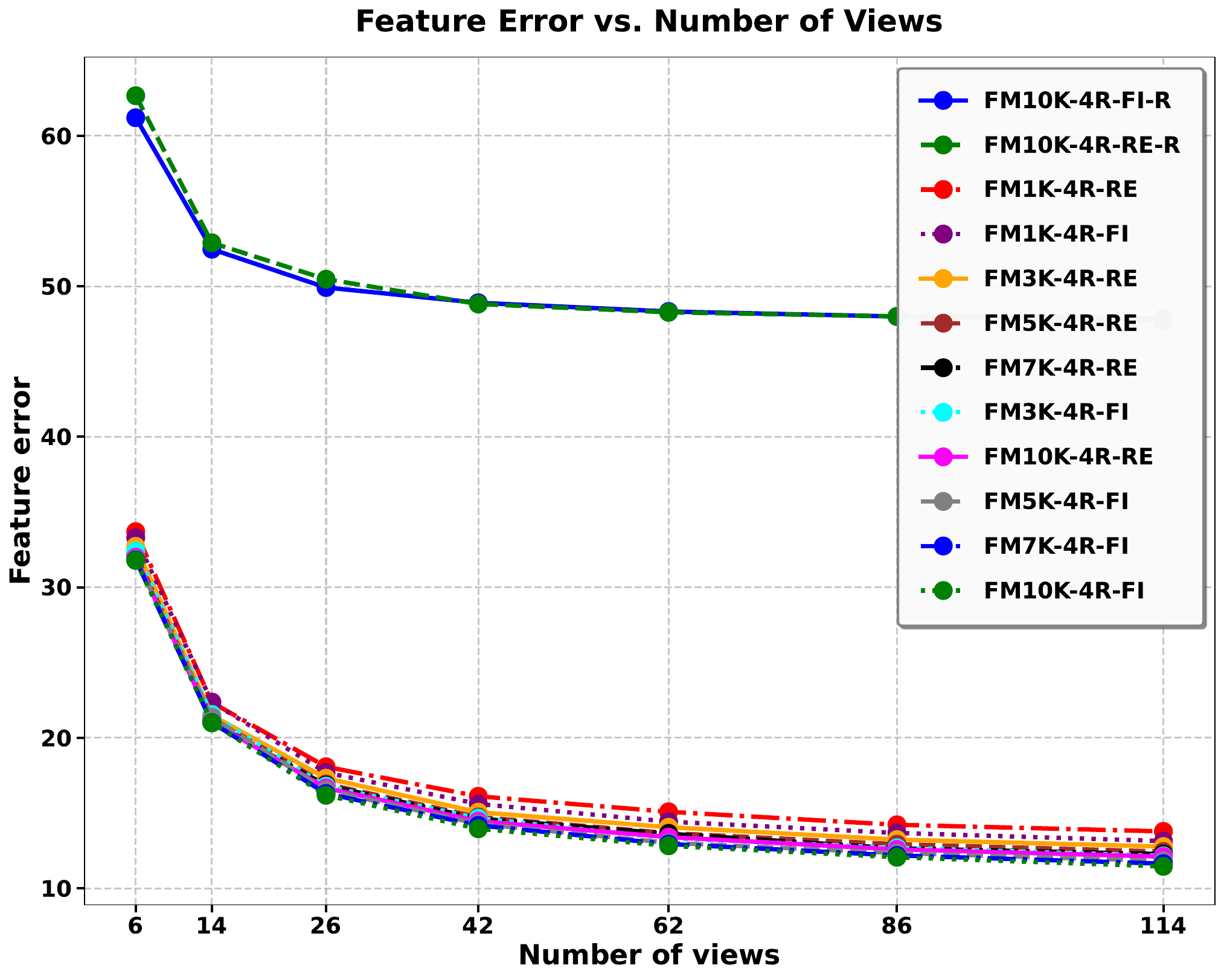}}}\\
(a)&(b)
\end{tabular}
\caption{(a) Uses the original mesh (\textbf{NO Feature-mesh sampling}) with different configurations of the feature extraction algorithm. (b) Combines Feature-mesh sampling with different configurations. Legend: RM: Regular mesh with no sampling is used; FMXK: Feature-mesh sampling is used with X points sampled; 1R: one rotation at 0°; 2R: two rotations at 0° and 180°; 3R: three rotations at 0°, 90°, and 270°; 4R: four rotations at 0°, 90°, 180°, and 270°; T4: same image repeated four times without rotation; RE: Regular viewpoint sampling; FI: Fibonacci viewpoint sampling; R: Random pairing instead of the proposed one.}
\label{fig:Feature_Error}
\end{figure}

\subsection{Results and Analysis}
Figure \ref{fig:Feature_Error} presents the results of our feature invariance analysis. Each plot shows the feature discrepancy (E) as a function of the number of viewpoints, with lower values indicating better invariance to symmetry. The following observations are based on the said figure:

\vspace{0.1em}\noindent\textbf{Viewpoint Sampling}
As shown in both (a) and (b), increasing the number of viewpoints generally improves feature invariance. Moreover, the Fibonacci spiral sphere sampling strategy consistently outperformed regular sphere sampling but only by a small margin.

\vspace{0.1em}\noindent\textbf{Impact of Feature-mesh Sampling}
In (b) while using Feature-mesh sampling we observe that increasing the sampling density from 1K to 10K points does improve feature invariance across all viewpoint configurations and performs better than (a).

\vspace{0.1em}\noindent\textbf{Rotation Augmentation}
The results in (a) clearly demonstrate the benefit of rotation augmentation. Using 4 rotations (0°, 90°, 180°, 270°) yielded the lowest feature discrepancy compared to the rest. Importantly, the control condition T4 performed the same as using only one rotation augmentation, confirming that the improvement is due to increased viewpoint diversity rather than simply processing more images.

\vspace{0.1em}\noindent\textbf{Random vs. Symmetric Pairing}
For both (a) and (b) random pairing produce larger values overall. The results confirm that symmetric point pairs exhibit significantly lower feature distances than random pairs, providing strong empirical support for our symmetry detection approach.

\subsection{Connection to Symmetry Detection}
The strong invariance properties demonstrated in our analysis justifies our approach of identifying symmetric correspondences through feature similarity in Subsection \ref{Subsec:Symmetry_det}. Since symmetric counterparts exhibit similar features (as shown by our experiments), searching for nearest neighbors in feature space is an effective strategy for finding potential symmetric pairs. Furthermore, the optimal parameters identified (10K points, Fibonacci-sampled viewpoints, and 4 rotations) provide the ideal configuration for maximizing feature invariance.

These empirical findings validate our core hypothesis that foundation model features, when properly back-projected onto 3D geometry, capture symmetry-invariant properties that can be leveraged for effective symmetry detection.

\section{Experiments and Results}
\label{Sec:Experiments}

\subsection{Dataset and Metrics}
\vspace{0.1em}\noindent\textbf{Dataset}
\label{Subsec:Dataset}
In this study, we use the same dataset and methodology as PRS-Net~\cite{Gao2021} and E3Sym~\cite{Li2023}. These works employ the ShapeNet~\cite{ShapeNet} dataset to train unsupervised neural networks for planar symmetry detection. For evaluation, both methods use a test set consisting of 1000 objects with known symmetry planes.

Since our method does not require training, we do not utilize the ShapeNet training data. Instead, we directly process the objects from the evaluation set to detect symmetries.

The 1000 objects in the test set are pre-aligned with the 3D coordinate axes, meaning that their known symmetry planes generally coincide with the principal coordinate planes. To introduce additional variability, we generate an alternative version of the test set by applying random rotations to the objects before evaluation.

\vspace{0.1em}\noindent\textbf{Metrics}
\label{Subsec:Metrics}
The following metrics are the same ones presented in E3Sym~\cite{Li2023} and are used for the experiments because they represent effectively if a detected symmetry is close to the respective ground truth and if is in fact a symmetry.

\vspace{0.1em}\noindent\textbf{Symmetry Distance Error (SDE)}  
Given a detected symmetry plane $T$ and an object $O$, this metric measures the geometric discrepancy between $O$ and its reflected counterpart $T(O)$. It is defined as:

\begin{equation}
    SDE(O, T) = \frac{1}{|O|} \sum_{o \in O} \min_{\hat{o} \in T(O)} \|o - \hat{o}\|_2^2
\end{equation}

Since the objects are triangular meshes, the minimum distance between a sampled point and the triangles of the reflected mesh is computed. For efficiency, a random subset of 1000 points is sampled per object. For a set of objects, the SDE is computed for each detected plane, and the average SDE across all detected planes is reported.

\vspace{0.1em}\noindent\textbf{F-score}  
The F-score metric evaluates the correctness and completeness of detected symmetry planes against ground truth. For a detected plane $T$, if a ground truth plane $\hat{T}$ exists such that their distance is below a threshold, $T$ is considered a true positive (TP). Otherwise, it is a false positive (FP). Any unmatched ground truth plane is a false negative (FN). The precision and recall are defined as $PR = \frac{TP}{TP + FP}, \ RE = \frac{TP}{TP + FN}$. The F-score is then given by the following formula:

\begin{equation}
    F\text{-}score = \frac{2 \cdot PR \cdot RE}{PR + RE}
\end{equation}

\noindent The distance between two planes $P = (a_P, b_P, c_P, d_P)$ and $Q = (a_Q, b_Q, c_Q, d_Q)$ accounts for the ambiguity in plane normal orientation $\text{dist}(P, Q) = \min\left(\|P - Q\|, \|P + Q\|\right)$. To account for different tolerance levels, the F-score is averaged over the following thresholds $[0.05, 0.1, 0.15, 0.2]$.

\subsection{Comparison with State-of-the-Art}

We evaluated our method against a comprehensive set of existing approaches for 3D symmetry detection, including both traditional geometry-based and recent learning-based approaches. Table~\ref{tab:Results_Ours} presents the quantitative results on the ShapeNet subset and metrics, both defined in~\ref{Subsec:Metrics}.

\begin{table}[ht]
\centering
\resizebox{\textwidth}{!}{
\begin{tabular}{|l|ccccccccccccc|}
\hline
\multirow{2}{*}{\textbf{Method}}  & PCA & OBB & RSD & ADS & PASD & PRST & PRST+GEDT & PAS & PRS-Net & E3Sym & Diff. & Self-Sym & Ours \\
&  & ~\cite{Chang2011} & ~\cite{Kazhdan2002} & ~\cite{Martinet2006} & ~\cite{Mitra2006} & ~\cite{Podolak2006} & ~\cite{Podolak2006} & ~\cite{Korman2015} & ~\cite{Gao2021} & ~\cite{Li2023} & ~\cite{Sipiran2014}  & ~\cite{Aguirre2025} &  \\
\hline
$\mathbf{SDE (\times 10^{-4})}$ & 3.32 & 1.25 & 0.90 & 3.95 & 14.2 & 1.78 & 1.60 & 1.75 & 0.86 & 0.46 & 7.10 & 4.72 & \textbf{0.31} \\
\textbf{F-score}               & 0.692 & 0.740 & 0.684 & 0.694 & 0.322 & 0.619 & 0.646 & 0.678 & 0.712 & 0.753 & 0.460 & 0.700 & \textbf{0.865} \\
\hline
\end{tabular}
}
\caption{Quantitative comparison of symmetry detection methods on the ShapeNet subset for SDE ($\times 10^{-4}$) and F-score (\%). Methods from PCA up to E3Sym are from the E3Sym paper~\cite{Li2023}, the next two (Diffusion and Self-Sym) are from Self-Sym~\cite{Aguirre2025}, and the final is our method, which shows an improvement of $\sim15\%$ in F-score.}
\label{tab:Results_Ours}
\end{table}

Furthermore, we perform the same experiment with different parameters to analyze the performance of different configurations for our method. Table~\ref{tab:Results_ablation} show the results for this experiment, the best result is used for Table~\ref{tab:Results_Ours}.

\begin{table}[ht]
\centering
\resizebox{\textwidth}{!}{
\begin{tabular}{|l|cccc|cccc|cccc|cccc|}
\hline
\multirow{2}{*}{\textbf{Method and views}} 
& \multicolumn{4}{c|}{RM-1R-RE} 
& \multicolumn{4}{c|}{RM-4R-FI} 
& \multicolumn{4}{c|}{FM10K-1R-RE} 
& \multicolumn{4}{c|}{FM10K-4R-FI} \\
\cline{2-17} 
 & 6 & 42 & 86 & 114 & 6 & 42 & 86 & 114 & 6 & 42 & 86 & 114 & 6 & 42 & 86 & 114 \\
\hline
$\mathbf{SDE (\times 10^{-4})}$ & 1.90 & 1.76 & 1.74 & 1.83 & 1.83 & 1.76 & 1.68 & 1.75
                                & 0.31 & 0.30 & 0.31 & 0.31 & 0.31 & \textbf{0.31} & \textbf{0.31} & 0.30\\
\textbf{F-score}               & 0.798 & 0.800 & 0.799 & 0.796 & 0.800 & 0.809 & 0.813 & 0.812
                                & 0.864 & 0.862 & 0.864 & 0.863 & 0.864 & \textbf{0.865} & \textbf{0.865} & 0.864\\
\hline
\end{tabular}
}
\caption{Results of different configurations by our method using views (6, 42, 86, 114) for SDE ($\times 10^{-4}$) and F-score (\%). Method names are given by the following: RM: Regular mesh with no sampling; FM10K: Feature-mesh sampling with 10k points sampled; 1R: one rotation at 0°; 4R: four rotations at 0°, 90°, 180°, 270°; RE: Regular viewpoint sampling; FI: Fibonacci viewpoint sampling.}
\label{tab:Results_ablation}
\end{table}

\section{Discussion}
\label{Sec:Discussion}
The results presented in Table~\ref{tab:Results_ablation} are relatively consistent across configurations. We attribute this to the design of the proposed symmetry algorithm ~\ref{Subsec:Symmetry_det} which is robust in that it forms trios of points rather than relying solely on pairwise matching. This approach generates additional candidate symmetry planes, enhancing the method’s performance even with the worst-performing configuration used in the invariance analysis. With respect to Table~\ref{tab:Results_Ours}, our method achieves the lowest Symmetry Distance Error (SDE), indicating highly precise and accurate symmetry detection. Additionally, it obtains the highest F-score, demonstrating strong generalization capabilities given that is training-free. And could prove useful in real world applications.

\noindent This proposed methodology could be further improved in the future to encompass more than only global symmetries, such as axial symmetries or partial objects. Furthermore, our implementation currently supports only triangular mesh inputs and adapting the approach to point clouds could increase its utility in practical scenarios.

\noindent \textbf{Our implementation, evaluation code, and datasets will be released upon publication.}

\section{Conclusions}
\label{Sec:Conclusions}
We presented a novel training-free, zero-shot approach for 3D symmetry detection that leverages visual features from foundation vision models. By back-projecting features extracted from rendered images to the 3D geometry, our method effectively identifies symmetry planes without requiring any training data.
Future work could explore extending our method to other types of symmetries, and applying similar principles to other geometric analysis tasks.
\newpage

\bibliographystyle{unsrt}  


\newpage
\section*{A Evaluation of hybrid method}

In Related Work, we mentioned Langevin~\cite{Langevin2024} as a hybrid approach to symmetry detection. We attempted to evaluate this method on our test set for fair comparison. However, due to insufficient parameter specification in the original paper, regarding clustering parameters, and we could not achieve competitive results to show in Table~\ref{tab:Results_Ours}.

Additionally, we found the method to be particularly slow (computationally expensive), which limited our evaluation to only 350 of the 1000 objects defined in Section~\ref{Subsec:Dataset}. Using clustering values of $eps=0.02$ and $MinPts=15$, we obtained an F-score of 0.2476 and SDE($\times10^{-4}$) of 7.17. The SDE results are relatively good, indicating that when the method detected planes, they were geometrically accurate. However, the low F-score suggests issues with either missing true symmetry planes or detecting false positives.

Both methods, Langevin and ours are hybrid-like, so we would like a fair comparison in the future.

\section*{B Visual guidance of the processes}
\begin{figure}[ht]
\centering
\begin{tabular}{cc}
\hbox{\fbox{\includegraphics[width=0.45\linewidth]{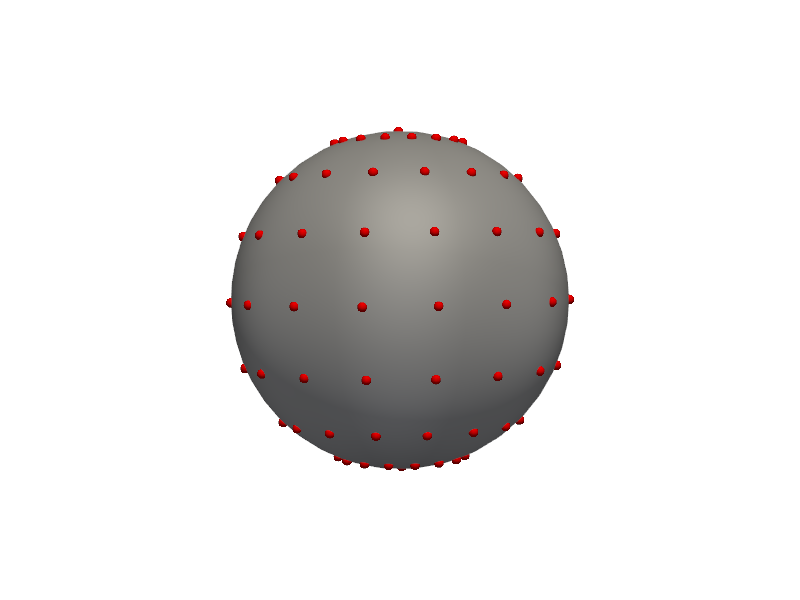}}}&
\hbox{\fbox{\includegraphics[width=0.45\linewidth]{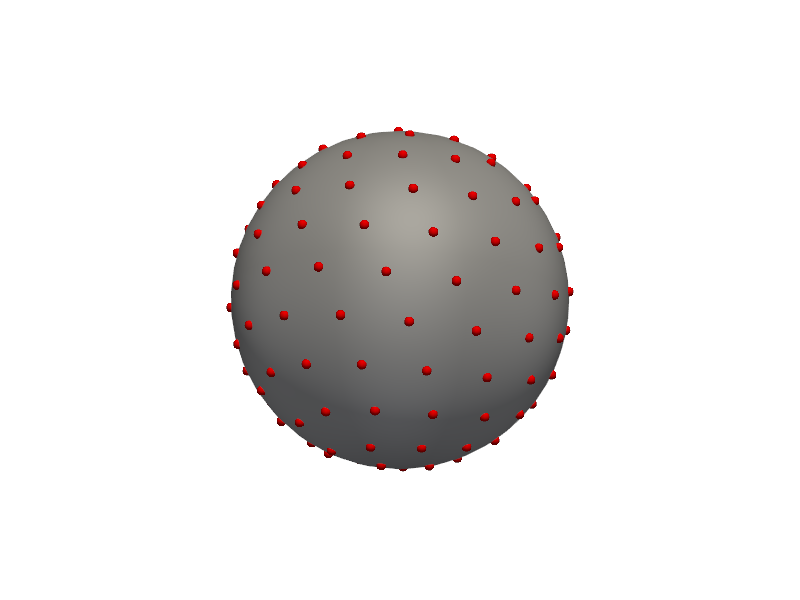}}}\\
(a)&(b)
\end{tabular}
\caption{(a) Regular sphere sampling, (b) Fibonacci sphere sampling.}
\label{fig:sphere_sampling}
\end{figure}

Figure \ref{fig:sphere_sampling} shows both regular and Fibonacci sphere sampling.

\begin{figure}[ht]
\begin{center}
\fbox{\includegraphics[width=.95\linewidth]{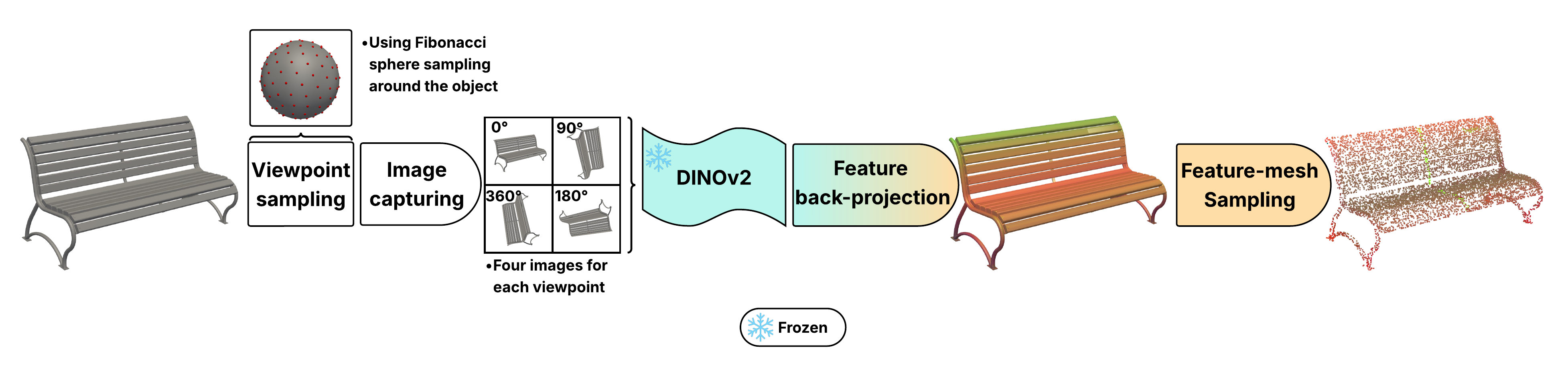}}
\end{center}
\caption{Visual representation of the feature backprojection process.}
\label{fig:backprojection}
\end{figure}

Figure \ref{fig:backprojection} shows the process of feature backprojection.

\begin{figure}[ht]
\begin{center}
\fbox{\includegraphics[width=.95\linewidth]{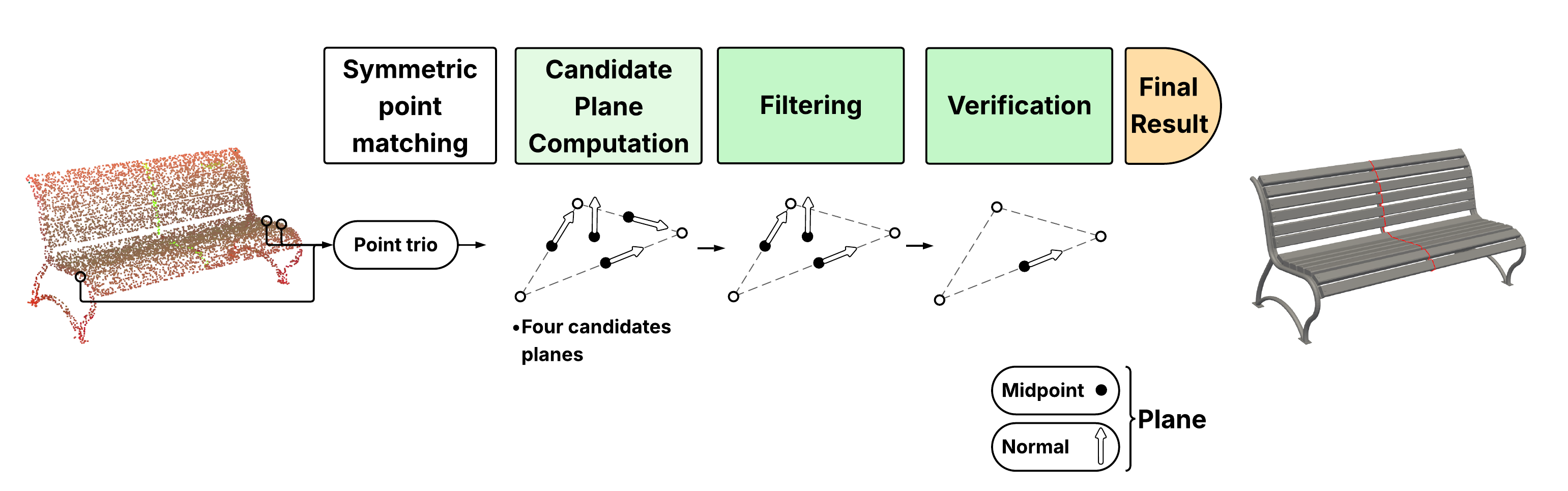}}
\end{center}
\caption{Visual representation of the proposed symmetry algorithm.}
\label{fig:symmetry_comp}
\end{figure}

Figure \ref{fig:symmetry_comp} shows the proposed symmetry algorithm.

\section*{C More visual examples of our method}

\begin{figure}[ht]
\begin{center}
\fbox{\includegraphics[width=.95\linewidth]{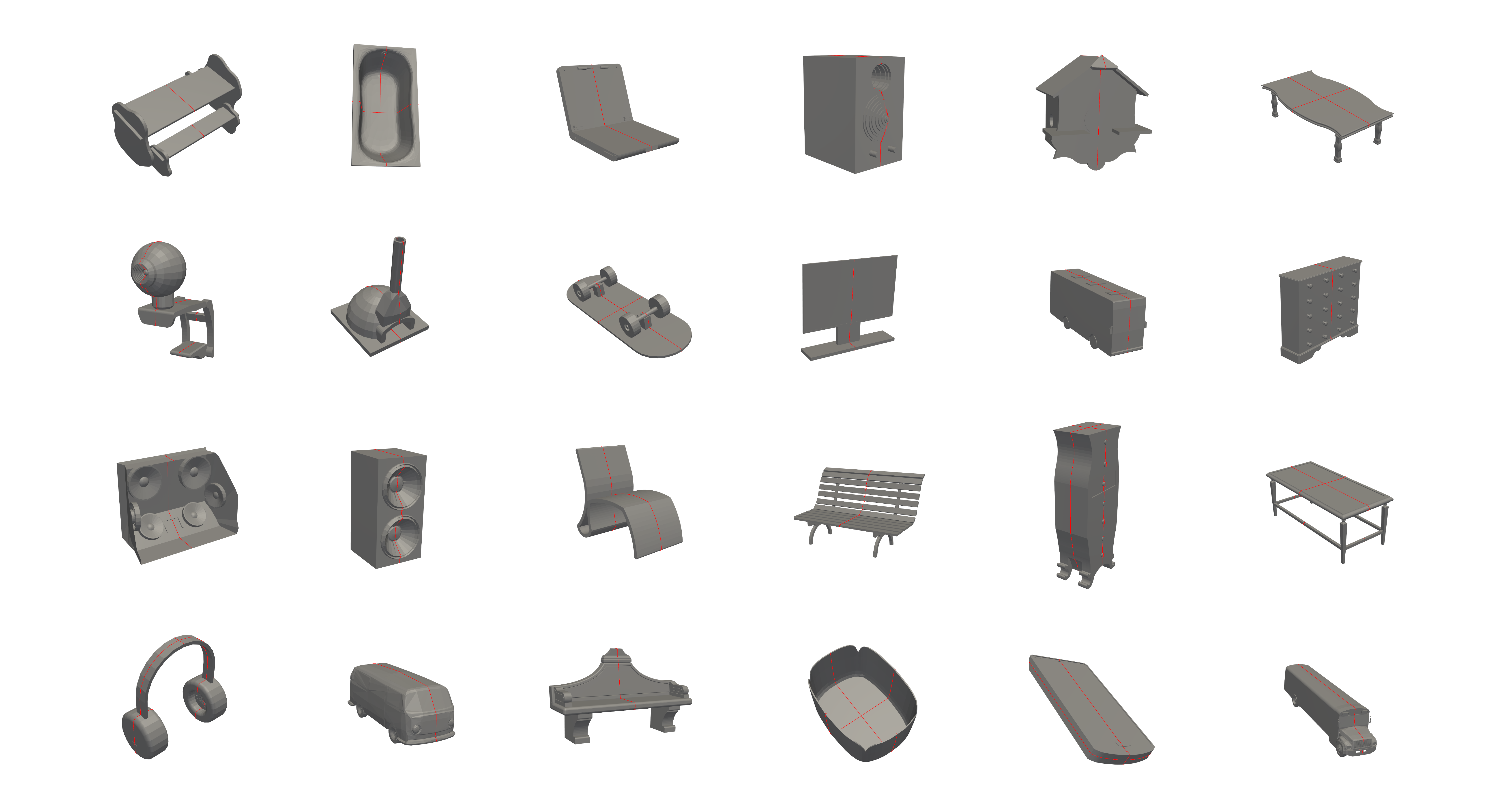}}
\end{center}
\caption{More 3D objects and symmetry planes detected by our method. Symmetry planes are outlined in red.}
\label{fig:more_examples}
\end{figure}

Figure \ref{fig:more_examples} shows more examples of symmetry planes detected by our method.

\end{document}